\documentclass[a4paper,twoside]{article}

\usepackage{epsfig}
\usepackage{subcaption}
\usepackage{calc}
\usepackage{amssymb}
\usepackage{amstext}
\usepackage{amsmath}
\usepackage{amsthm}
\usepackage{multicol}
\usepackage{pslatex}
\usepackage{apalike}
\usepackage{algorithm2e}
\usepackage{booktabs}
\usepackage{multirow}
\usepackage[bottom]{footmisc}
\usepackage{SCITEPRESS}     
\usepackage{url}
\input{mysymbol.sty}

\begin{document}

\title{GNNs for Time Series Anomaly Detection: An Open-Source Framework and a Critical Evaluation}

\author{\authorname{Federico Bello\sup{1}, Gonzalo Chiarlone\sup{1}\sup{2}, Marcelo Fiori\sup{1}\sup{3}\orcidAuthor{0000-0002-3732-1778}, Gast\'on Garc\'ia Gonz\'alez\sup{1}\orcidAuthor{0009-0002-6652-7713} and Federico Larroca\sup{1}\sup{3}\orcidAuthor{0000-0001-7893-2201}}
\affiliation{\sup{1}Facultad de Ingenier\'ia, Universidad de la Rep\'ublica, Montevideo, Uruguay}
\affiliation{\sup{2}Pento, Montevideo, Uruguay}
\affiliation{\sup{3}Centro Interdisciplinario en Ciencia de Datos y Aprendizaje Automático (CICADA), Universidad de la Rep\'ublica, Uruguay}
\email{\{federico.bello, gonzalo.chiarlone,mfiori,gastong,flarroca\}@fing.edu.uy}
}


\abstract{There is growing interest in applying graph-based methods to Time Series Anomaly Detection (TSAD), particularly Graph Neural Networks (GNNs), as they naturally model dependencies among multivariate signals. GNNs are typically used as backbones in score-based TSAD pipelines, where anomalies are identified through reconstruction or prediction errors followed by thresholding.
However, and despite promising results, the field still lacks standardized frameworks for evaluation and suffers from persistent issues with metric design and interpretation. 
We thus present an open-source framework for TSAD using GNNs, designed to support reproducible experimentation across datasets, graph structures, and evaluation strategies. Built with flexibility and extensibility in mind, the framework facilitates systematic comparisons between TSAD models and enables in-depth analysis of performance and interpretability.
Using this tool, we evaluate several GNN-based architectures alongside baseline models across two real-world datasets with contrasting structural characteristics. 
Our results show that GNNs not only improve detection performance but also offer significant gains in interpretability, an especially valuable feature for practical diagnosis. 
We also find that attention-based GNNs offer robustness when graph structure is uncertain or inferred.
In addition, we reflect on common evaluation practices in TSAD, showing how certain metrics and thresholding strategies can obscure meaningful comparisons. 
Overall, this work contributes both practical tools and critical insights to advance the development and evaluation of graph-based TSAD systems.}

\onecolumn \maketitle \normalsize \setcounter{footnote}{0} \vfill







\section{\uppercase{Introduction}}




%


Anomaly detection plays a central role in domains such as fraud detection~\cite{Hilal2022}, cybersecurity~\cite{Siddiqui2019}, industrial monitoring~\cite{nizam2022real}, and medical diagnostics~\cite{Spence}. Within this broad area, Time Series Anomaly Detection (TSAD) focuses on identifying unexpected behaviors in temporally ordered data~\cite{shaukat2021tsad}. In recent years, and driven by its success in other domains, Deep Learning (DL) has been increasingly applied to TSAD~\cite{zamanzadeh2024deep}. 


The typical pipeline for anomaly detection using deep learning consists of two key components: a backbone model and a scoring module~\cite{Jin2023}. The backbone is trained under the assumption that most data is normal, and the scoring module flags deviations via reconstruction or prediction errors, serving as proxies for identifying unexpected patterns. However, standard DL models often treat multivariate time series as sequences of independent feature vectors, neglecting structural dependencies that may be essential for accurate and interpretable detection.


Graph Neural Networks (GNNs), designed to operate on graph-structured data, have shown promise for modeling such dependencies. By representing time series as graphs, GNNs enable the joint modeling of temporal dynamics and inter-variable dependencies through message passing, effectively capturing complex relational structures~\cite{Chen_2022,Deng2021GDN}. 
This ability has spurred a growing interest in graph-based TSAD~\cite{Jin2023}, where GNNs act as backbones for reconstruction- or prediction-based anomaly scoring.
Yet, despite promising performance, the field remains fragmented: implementations are rarely comparable, evaluation practices vary widely, and metric design often leads to inconsistent or misleading conclusions. As a result, progress is difficult to quantify and reproduce.

To address these issues, we introduce a unified, modular, and open-source framework for graph-based TSAD.\footnote{The source code and configuration files for our framework are available at \url{https://github.com/GraGODs/GraGOD}.}  Built in PyTorch with reproducibility and extensibility in mind, the framework provides standardized procedures for data handling, model configuration, and evaluation. It natively supports both graph-based and non-graph-based approaches, enabling fair comparisons across modeling paradigms. Crucially, it integrates a diverse set of evaluation metrics, from classical point-wise precision and recall to range-based~\cite{lee2018range,tatbul2018precision} and threshold-agnostic measures such as the Volume Under Surface (VUS)~\cite{paparrizos2022VUS}, offering a consistent environment for methodological analysis.




Using our framework, we conduct a systematic comparative study of representative GNN-based methods and baselines across datasets with contrasting structural characteristics. The results reveal how graph topology, thresholding strategy, and metric design interact to influence performance and interpretability. In particular, we find that attention-based GNNs offer robustness to uncertainty in graph structure while improving interpretability by localizing anomalies to specific nodes. Conversely, we show that common evaluation practices, especially those relying solely on point-wise or threshold-dependent metrics, can obscure genuine model differences.

Beyond empirical benchmarking, this work contributes methodological insights into how graph-based representations and evaluation metrics shape the behavior of TSAD systems. The proposed framework establishes a reproducible foundation for future research in pattern recognition of time series over graphs, facilitating the development of more reliable and interpretable anomaly detection methods.

The rest of this paper is structured as follows.
Section \ref{sec:problem} formalizes the problem of time series anomaly detection and presents the models and datasets considered in this work. 
In Sec.\ \ref{sec:shortcoming} we discuss the main methodological challenges in TSAD evaluation, reviewing the limitations of conventional point-wise metrics (e.g.\ precision and recall), but also their range-based extensions which attempt to account for the temporal extent of anomalies.
This section also introduces the proposed framework and its design principles for reproducible experimentation.
Equipped with our framework, Sec.\ \ref{sec:experiments} presents and discusses the benchmark results obtained with different models, metrics, and graph topologies. 
Finally, Sec.\ \ref{sec:conclusions} concludes the article.

\section{Problem Statement, Methods and Datasets}\label{sec:problem}




Classic TSAD methods had been classified into taxonomies by several authors \cite{Darban,Paparrizos,blazquez}.
At the coarsest level there are some common groups (eventually overlapping) like Statistical-, Clustering-, Distance-, or Density-based, as well as Forecasting- or Reconstruction-based techniques, which are the focus of this work. 
Forecasting-based detection builds a model (statistical or machine-learning-based) to predict the next point in the time series. Anomaly scores are then obtained from the residual of the predicted and the real value, and points whose errors exceed a threshold are flagged.
Reconstruction-based detection trains an autoencoder, PCA, or matrix-factorization model to compress and then reconstruct windows of observations.  If a window  cannot be accurately reconstructed (i.e., its reconstruction error is high), the corresponding region is deemed anomalous.

Numerous methods leveraging GNNs for TSAD have been proposed in recent years. For instance, more than thirty such works are discussed in the comprehensive review by \cite{Jin2023}, to which we refer the reader for further details.
Across these GNN-based approaches for multivariate TSAD, researchers have explored a diverse set of modeling tools to capture spatiotemporal dependencies and distinguish normal from anomalous behavior. Some methods \cite{Deng2021GDN,mtadGAT} learn explicit dependency graphs among variables using attention mechanisms to perform predictive or reconstructive modeling, thereby offering interpretable relations between sensors. Others \cite{GANF,VGCRN} adopt probabilistic formulations, leveraging normalizing flows for likelihood-based anomaly scoring or combining variational inference with graph convolution and recurrent units to model uncertainty and temporal dynamics. Another family \cite{GReLeN,FuSAGNet} focuses on relational or sparse graph learning, embedding graph-structure discovery directly within autoencoder or forecasting architectures to capture hidden dependencies. We now briefly present how the TSAD problem is formulated in this context, and describe the two state-of-the-art methods of the first family, included in this framework.

Let $\mathcal{X}$ be a set of $N \in \mathbb{N}^*$ distinct time series. We will denote a given time series as $\mathbf{X}_i = [x^{(1)}_{i}, x^{(2)}_{i}, \ldots, x^{(T)}_{i}]$, where $x^{(t)}_{i} \in \mathbb{R}$, and $T$ is the length of the time series. Depending on the application or the labeling of the data, the goal is to detect either an anomaly at a specific time series $i\in [1\ldots N]$, or an anomaly at a global level, meaning that the system presents an anomaly at a certain time.

The time series may inherently be connected through an underlying graph structure, which may be explicit, as is common in scenarios like sensor networks, or implicit, like causal dependencies in financial markets. In the latter case, a key assumption is that there exists some kind of correlation or dependencies between some of the time series, and therefore a graph can (must) be inferred. In this work we use datasets from both scenarios, presented below.
Therefore, given the multivariate set of $N$ time series, we will consider a graph $G$ with $N$ nodes, each of which correspond to a certain time series. The structure of the graph (i.e., the edges and their weights) plays a fundamental role. As mentioned, this structure may come beforehand from the problem itself, like an industrial pipeline with sensors, or may be abstract and learned from the data.


The experimental framework {compares} multiple GNN-based models, which produce their outputs by operating on the graph through message passing, alongside a structure-agnostic model serving as a benchmark. Each model is trained uniformly, functioning either as a forecaster or reconstructor. The models receive as an input a window of datapoints of size $w$, defined as:
\begin{equation}
\mathbf{X}^{(t)} = [{\mathbf{x}^{(t-w+1)}, \mathbf{x}^{(t-w+2)}, \ldots, \mathbf{x}^{(t)}}] \in \mathbb{R}^{N \times w},
\end{equation}
where each $\mathbf{x}^{{(\tau)}}$ is composed of the $\tau$-th datapoint of each time series, i.e. $\mathbf{x}^{(\tau)} = (x^{(\tau)}_1, x^{(\tau)}_2, \ldots, x^{(\tau)}_{N})$, and produce either an estimate $\hat{\mathbf{x}}^{(t+1)}$ (forecaster) or $\hat{\mathbf{X}}^{(t)}$ (reconstructor).

{The approaches based on GNNs compute these outputs by combining the time series using different architectures and supporting graphs. As previously mentioned, the graph can either be learned from the data (e.g.\ through correlations) or provided by the user. For example, assume we are using a forecasting-based method and we have an actual network $G$ connecting the nodes, which we will represent through the adjacency matrix $\bbA$ (which may include weights). Then, the trained GNN-based forecaster is a function $\bbPhi(\bbA,\bbtheta,\bbX^{(t)})=\hat{\bbx}^{(t+1)}$}, where each node has an associated $w$-dimensional signal as the input, and a 1-dimensional signal as the output. Note that $\bbA$ is fixed throughout all values of $t$ even if the architecture uses attention mechanisms. Furthermore, if we estimate/infer the graph (and thus the adjacency matrix), we perform this estimation once, meaning that $\hat{\bbA}$ is also fixed for all values of $t$ (see the discussion in Sec.\ \ref{topology}).


Given a forecasting- or reconstruction-based method, anomaly scores computed as prediction/reconstruction errors are used to flag an anomaly when it is above a certain threshold. This anomaly may be at the node level (i.e., large errors in predicting/reconstructing an individual time series) or graph level (i.e.\ considering the error in all time series). 

\noindent \textbf{Methods.} 
Currently, our experimental framework incorporates four distinct models. Firstly, a structure-agnostic Gated Recurrent Unit (GRU) and a custom-designed Graph Convolutional Network (GCN), both serving as baselines for comparative analysis. The GCN operates on a fixed given graph structure, and the anomaly scores are computed as forecasting errors.

We have also included two state-of-the-art GNN-based models, which we now briefly describe. The \textit{Graph Deviation Network} (GDN) \cite{Deng2021GDN} is a deep learning-based approach that learns the structure of dependencies between variables, and using this graph and an attention mechanism produces a prediction of the next value.
The \textit{Multivariate Time-series Anomaly Detection via Graph Attention Network} (MTAD-GAT)~\cite{mtadGAT}, similar to GDN, also uses a GNN approach. The key of this model is using two different GATs, \textit{feature oriented GAT} and \textit{time oriented GAT}, to map the relationships both between the features and the temporal dependencies. The training involves both a reconstruction and a forecasting model at the same time. The anomaly score is then computed as a combination of both the forecast and reconstruction errors.

\noindent \textbf{Datasets.} To evaluate these methods we have chosen two representative datasets: the TELCO and SWaT datasets.
The TELCO dataset \cite{dc-vae}, consists of twelve distinct time series. Each one represent common metrics tracked by a mobile internet service provider (normalized and anonymized), such as the quantity and value of prepaid data transfer fees, the number and cost of calls, the volume of data traffic, and additional related data.
The dataset spans seven months, divided into three months for training, one month for validation, and three months for testing. The most noticeable aspect is the significant imbalance between anomaly and normal data within the dataset, a common characteristic in anomaly detection datasets. 
A key element is the absence of an explicit graph structure. While the time series data may exhibit correlations, there is no physical structure or defined relationship connecting the series.

The SWaT (Secure Water Treatment) \cite{swat} dataset is a widely used benchmark for evaluating anomaly detection methods. It consists of time series data collected from a scaled-down six-stage water treatment plant that replicates real-world industrial control systems. The dataset includes 51 physical and network-related features and spans eleven days, where the initial seven days record normal system operations, and the subsequent four days contain both normal and attack scenarios. The attacks were intentionally introduced and encompass both cyber and physical threats to the system. Notably, the SWaT dataset exhibits an inherent relational structure, as sensors within the same treatment stage often measure correlated physical properties like flow rate, pressure, or water level. 
An undirected graph is constructed where an edge is added between two nodes if the sensors measure similar properties or are in the same stage of the process.

Naturally, both TELCO and SWaT have limitations, including mislabeled anomalies, distribution shifts, and run-to-failure bias, that affect model evaluation. Despite these challenges, they can serve as useful benchmarks when paired with qualitative analysis. Effective preprocessing, such as removing redundancies, fixing labels, and handling distribution shifts, is essential for reliable and fair model comparisons.

\section{Challenges in current TSAD methodologies}\label{sec:shortcoming}







\noindent \textbf{Point-wise Metrics.} Despite the growing interest in TSAD, the evaluation of model performance remains a challenging and often overlooked aspect. Many existing works still rely on Precision, Recall (sensitivity) and F1-score, based on the classification of each time point as either normal or anomalous. However, these point-wise metrics present significant limitations, as they fail to capture the sequential nature and typical range-based structure of anomalies in time series \cite{tatbul2018precision}. 

\begin{figure}
\centering
\includegraphics[width=0.45\textwidth]{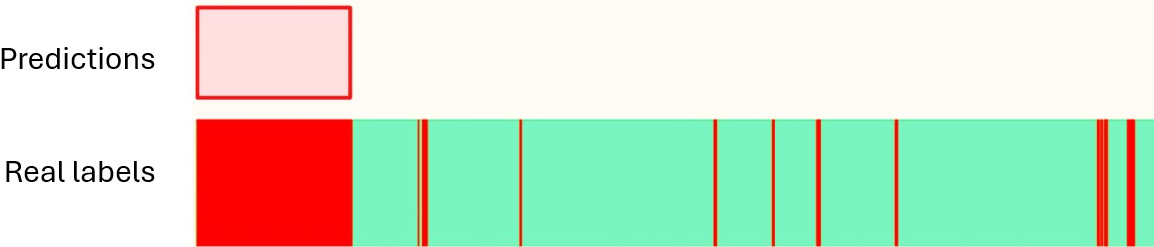}
\caption{Example of point-wise evaluation limitations. Although only one long anomaly is detected, point-wise metrics report a high Recall (0.8) and perfect Precision (1.0). This gives the false impression of good performance, despite the fact that most anomaly ranges in the dataset remain undetected.}\label{fig:point-wise_detection}
\end{figure}

In practice, it is often more important to detect as many distinct anomaly ranges as possible, even if their exact boundaries are missed, since identifying the occurrence of each anomaly is typically more valuable than precisely locating every anomalous point. For instance, consider the example in Fig.\ \ref{fig:point-wise_detection}, based on the SWaT dataset which presents a very similar pattern with a single long anomaly and several short ones. In this example, only the long anomaly is correctly detected. Despite this, point-wise metrics report a Recall of 0.8 and a Precision of 1.0, suggesting high performance. In reality, however, the model misses the majority of the anomaly ranges in the dataset, highlighting a major shortcoming of these metrics.

\noindent \textbf{Range-based metrics.} 
To address these limitations, range-based variants of Precision, Recall, and F1-score (denoted $P_T$, $R_T$ and $F1_T$ respectively in the sequel) have been proposed~\cite{tatbul2018precision}. These metrics account for the temporal extent of anomalies by rewarding partial overlap, penalizing fragmentation, and weighting detections by positional relevance, offering a more faithful evaluation than point-wise measures.

\begin{figure}
\centering
\includegraphics[width=0.45\textwidth]{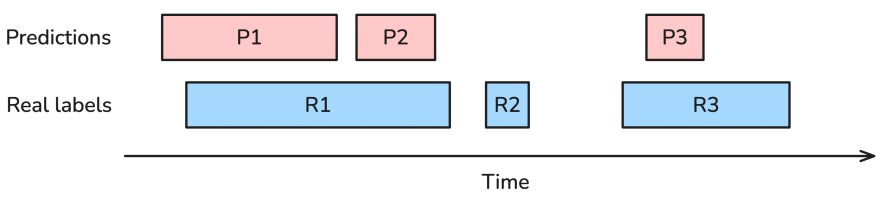}
\caption{
Range-based recall example. 
R1 and R3 get a high existence reward, R2 none. R1 obtains a high size score, R3 a low one, and R2 none. Cardinality is high for R3 but lower for R1, since the anomaly is detected as two separate segments instead of one. No position reward is considered here.
}\label{fig:range_anomaly}
\end{figure}

For example, range-based Recall $R_T$ evaluates how effectively a detector identifies true anomalous intervals by checking whether an anomaly is detected at all, even if partially (existence reward); how much of its duration is correctly identified (size); which parts are detected, e.g., if early detection is more critical (position); and whether it is reported as a single continuous range or split into fragments (cardinality). An illustrative example is shown in Fig.~\ref{fig:range_anomaly}.


These metrics require careful configuration, as their (several) parameters strongly influence results, and poor choices can lead to misleading conclusions. For example, in datasets with long anomalies, neglecting the cardinality component may allow multiple overlapping predictions on the same anomaly to artificially boost precision.

\begin{figure}
\centering
\includegraphics[width=0.45\textwidth]{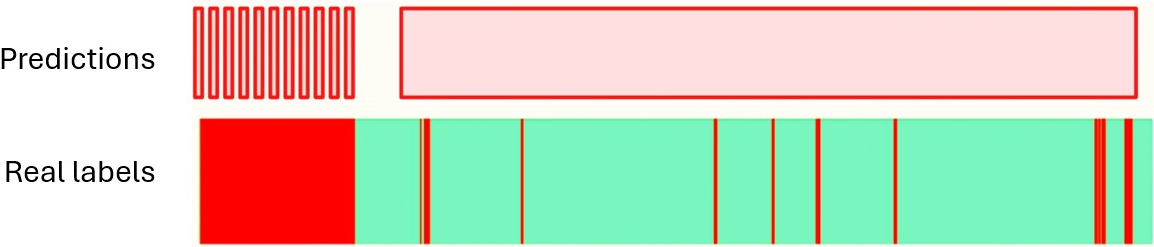}
\caption{Example of how range-based metrics' configuration can misrepresent performance. The model produces an overall poor prediction: the long anomaly is detected through multiple fragmented predictions, and several extended false positives occur across the timeline. However, under certain range-based metric configurations, such as existence-only recall and precision without cardinality penalty, the evaluation yields a high performance, masking the model's true shortcomings.}\label{fig:range_detection}
\end{figure}

Figure~\ref{fig:range_detection} illustrates this risk. Here, a long anomaly is detected through several fragmented predictions, with an additional long (and mostly false) detection across the rest of the timeline. Yet, if range-based Recall is configured to reward only the existence of overlap, the score reaches 1.0, since every anomaly is at least partially detected. Likewise, neglecting the cardinality penalty in range-based Precision allows the multiple predictions within the long anomaly to counterbalance the extended false positive, producing a value close to 1.0 despite poor detection quality.


\noindent \textbf{Threshold-agnostic Metrics.} 
To avoid the dependence on a specific value of the threshold, we will use the recently proposed Volume Under Surface (VUS), in its VUS-ROC and VUS-PR variants~\cite{paparrizos2022VUS}. 
This metric generalizes the traditional AUC concept from binary classification by integrating model performance over multiple buffer sizes around the annotated anomaly ranges. This allows for a continuous assessment of robustness to label imprecision and misalignment. VUS computes the volume under the surface generated by simultaneously varying both the decision threshold and the buffer parameter, eliminating dependence on specific hyperparameters or fixed thresholds. As a result, it provides a more robust and comprehensive evaluation for anomaly detection models.


\noindent \textbf{Lack of framework.} 
One of the most persistent challenges in TSAD research is the absence of a unified, standardized framework for systematically comparing detection methods. Existing implementations are typically tied to specific models, datasets, experimental setups, and metric choices—some of which, as discussed earlier, can produce misleading conclusions. This fragmentation limits reproducibility, constrains the scope of comparative studies, and makes it difficult to assess the real-world applicability of proposed models.


To address this gap, we introduce GraGOD, a modular and extensible open-source framework for the evaluation and comparison of machine learning and deep learning-based TSAD models. Unlike existing solutions~\cite{tsod-repo} that offer limited flexibility, GraGOD is designed as a collaborative, research-oriented framework where new models, datasets, and metrics can be seamlessly integrated. Its architecture natively supports both graph-based and non-graph-based methods, enabling fair and transparent comparison across different paradigms.

GraGOD provides a comprehensive experimental management system for TSAD, with a focus on GNN research. It supports end-to-end experimentation, including data preprocessing, model training, prediction, and hyperparameter tuning. The framework’s command-line interface allows users to orchestrate these processes, ensuring reproducibility and control over experiment configuration. GraGOD also integrates automated metric computation of all the metrics mentioned in this work, and provides modules for visualizing anomalies, helping researchers interpret model behaviors beyond numerical scores.

From a development perspective, GraGOD enforces a consistent project structure for code organization, dataset management, and result logging. It supports iterative experimentation and scalable execution, making it suitable for large-scale data analysis and computationally intensive model tuning. The framework’s design encourages community contributions by simplifying the addition of new datasets, models, and evaluation metrics through a well-documented API. Ultimately, GraGOD establishes a reproducible and extensible foundation for TSAD research, accelerating methodological progress and promoting transparent, comparable experimentation.

\section{Benchmark}\label{sec:experiments}






This section presents the experimental results, focusing on the impact of graph topology, threshold selection strategies, model interpretability, and the limitations of standard training paradigms. 
Training details can be consulted on the source code, although we highlight that thresholds (when used) were chosen to maximize the F1 score on the validation dataset, whereas the configuration of VUS metrics are left as suggested in the original paper~\cite{paparrizos2022VUS}. 

\subsection{Initial Benchmark Results}

Table~\ref{test_metrics} reports baseline results for the four anomaly detection models on both datasets. In this first experiment, all graph based models used a fully connected graph topology. A first observation is that the GDN, MTAD-GAT, and GRU models consistently achieve similar VUS scores, suggesting their predictive abilities are closely matched. It is interesting to note that VUS metrics are much lower for the TELCO dataset than for SWaT. This is indicative of poor separability between normal and anomalous scores. Since a random classifier yields a VUS-ROC of 0.5, the low VUS-ROC indicates that the models cannot easily distinguish anomalies.

\begin{table}
\centering
\caption{Test metrics for the TELCO and SWaT datasets using fully connected graphs.}
\resizebox{\columnwidth}{!}{
\begin{tabular}{@{}|l| l| c| c| c |c |c |c |c |c|@{}}
\toprule
\textbf{Dataset} & \textbf{Model} & $P$ & $R$ & $F1$ & $P_T$ & $R_T$ & $F1_T$ & $VUS$-$ROC$ & $VUS$-$PR$ \\
\midrule
\multirow{4}{*}{SWaT}
& GCN        & 0.80 & \textbf{0.79} & 0.80 & 0.06 & \textbf{0.33} & 0.10 & 0.72 & 0.55 \\
& GDN        & \textbf{1.00} & 0.75 & 0.85 & \textbf{1.00} & 0.07 & \textbf{0.13} & 0.85 & 0.73 \\
& MTAD-GAT & 0.00 & 0.00 & 0.00 & 0.00 & 0.00 & 0.00 & \textbf{0.88} & 0.74 \\
& GRU        & 0.98 & 0.76 & \textbf{0.86} & 0.08 & 0.24 & 0.12 & 0.86 & \textbf{0.77} \\
\midrule
\multirow{4}{*}{TELCO}
& GCN        & 0.15 & 0.10 & 0.08 & 0.09 & 0.29 & 0.11 & \textbf{0.64} & 0.05 \\
& GDN        & 0.32 & \textbf{0.18} & 0.11 & 0.30 & \textbf{0.48} & 0.25 & 0.62 & 0.08 \\
& MTAD-GAT & \textbf{0.39} & 0.16 & 0.10 & 0.34 & 0.44 & 0.25 & 0.61 & 0.07 \\
& GRU        & \textbf{0.39} & 0.14 & \textbf{0.12} & \textbf{0.35} & \textbf{0.48} & \textbf{0.30} & 0.58 & \textbf{0.09} \\
\bottomrule
\end{tabular}
}
\label{test_metrics}
\end{table}

Furthermore, it is important to highlight the clear mismatch between VUS and threshold-dependent metrics: high performance in the former does not necessarily translate into strong performance in the latter. As we discussed before, VUS aggregates performance over all possible thresholds, so it can remain high even if no single threshold yields satisfactory results. An extreme example is MTAD-GAT on SWaT, which achieves highly competitive VUS values yet produces no correct predictions when thresholded. This suggests a threshold selection issue rather than a fundamental model failure. Since the threshold was chosen to maximize $F1$ on the validation set, the problem arises from a shift in the score distribution between validation and test data. This is further confirmed when using more sophisticated selection methods, such as Otsu’s algorithm~\cite{OtsuThreshod2022}, which yields an excellent $F1=0.8$ but a disappointing $F1_T=0.07$. In contrast, a dynamic threshold based on the rolling mean and standard deviation of recent scores produces $F1=0.41$ and $F1_T=0.25$, which, although modest in absolute terms, are notably more consistent with each other than the results obtained with other methods. These findings underscore the critical role of both metric choice and robust thresholding strategies in TSAD evaluation.

\begin{figure}
    \centering
    \begin{minipage}[b]{0.49\linewidth}
    \centering
    \scriptsize{GCN}
        \includegraphics[width=\linewidth]{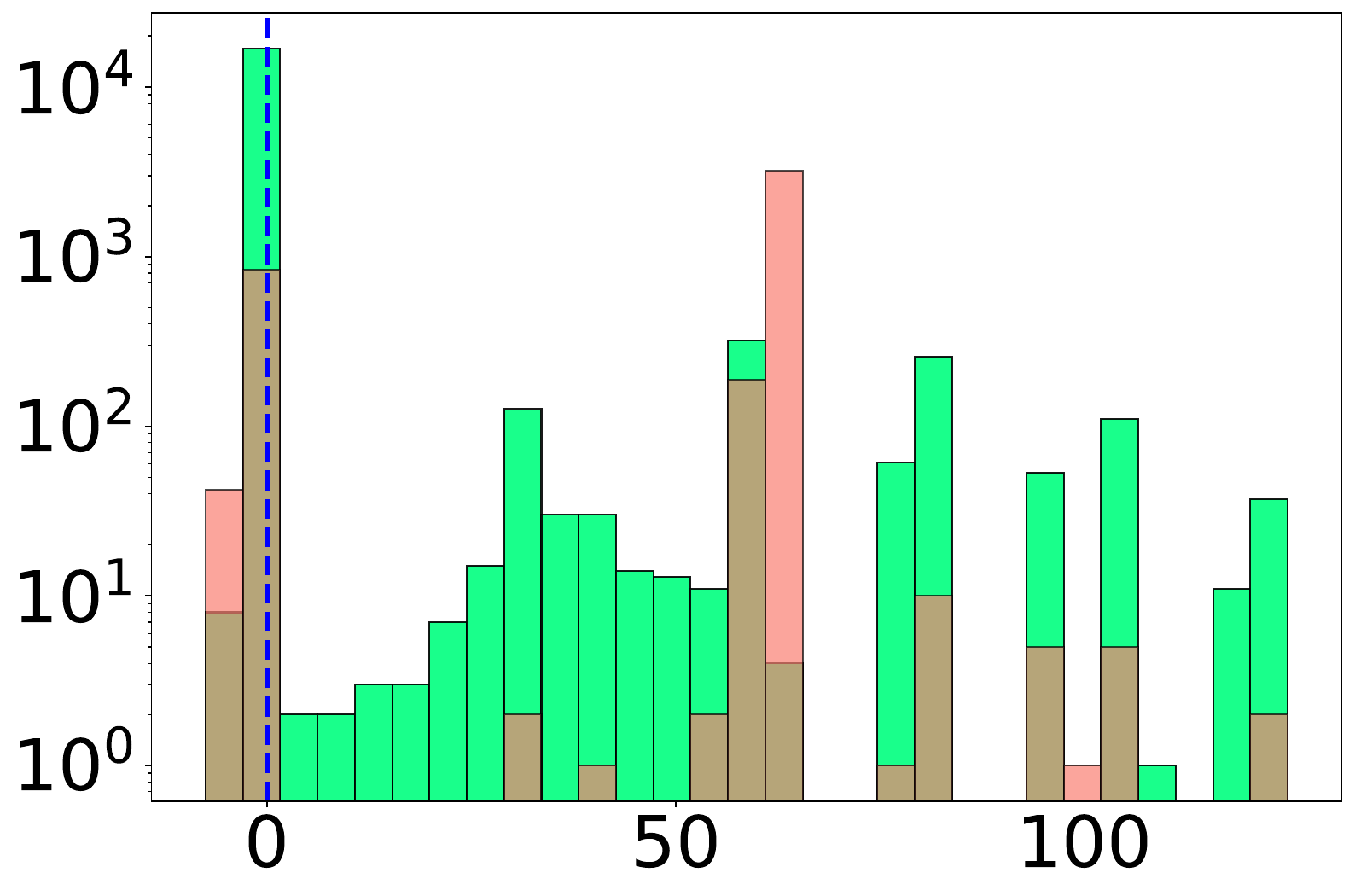}
    \end{minipage}
    \hfill
    \begin{minipage}[b]{0.49\linewidth}
    \centering
    \scriptsize{GDN}
        \includegraphics[width=\linewidth]{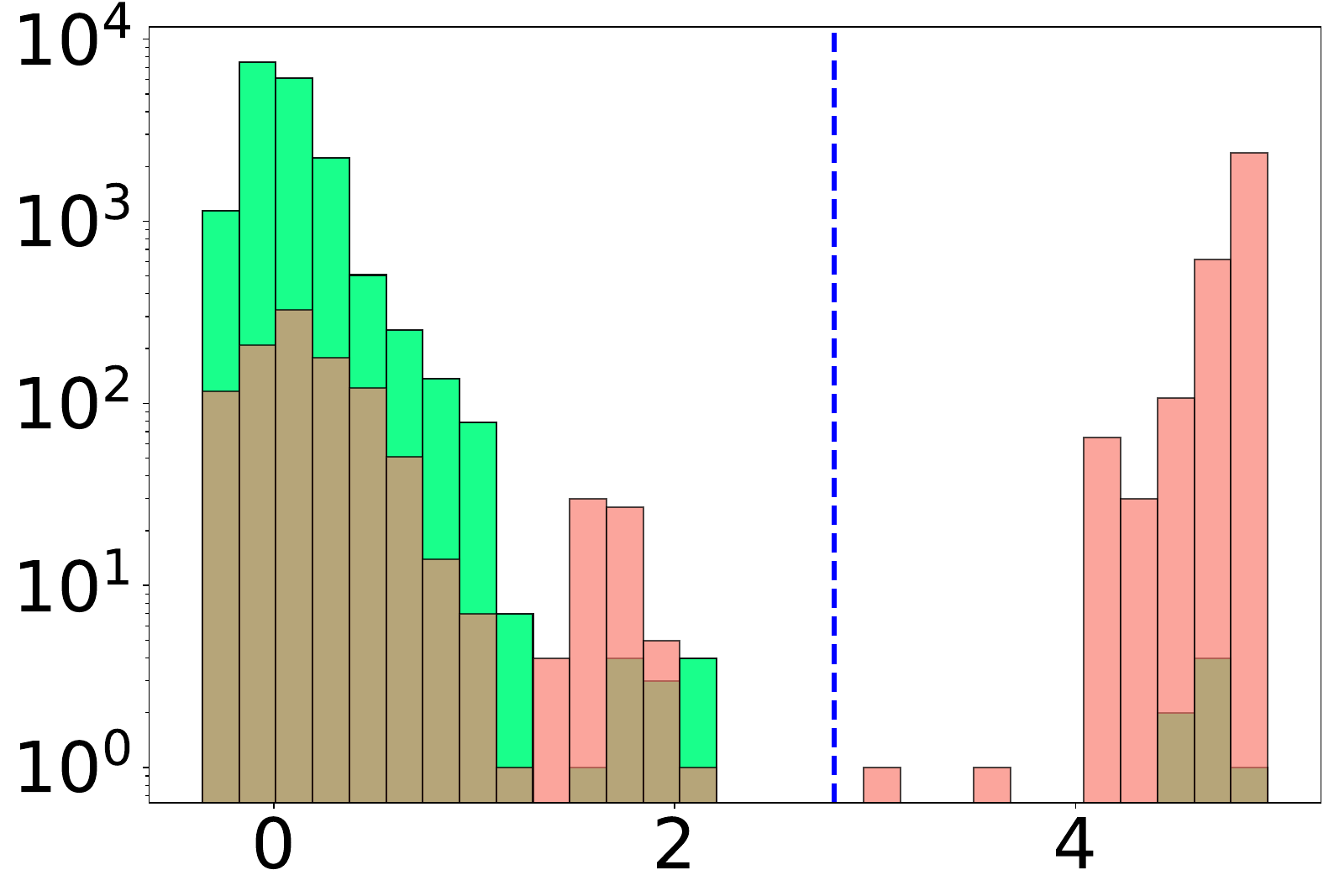}
    \end{minipage}
    \par\medskip
    \begin{minipage}[b]{0.49\linewidth}
    \centering
    \scriptsize{MTAD-GAT}
        \includegraphics[width=\linewidth]{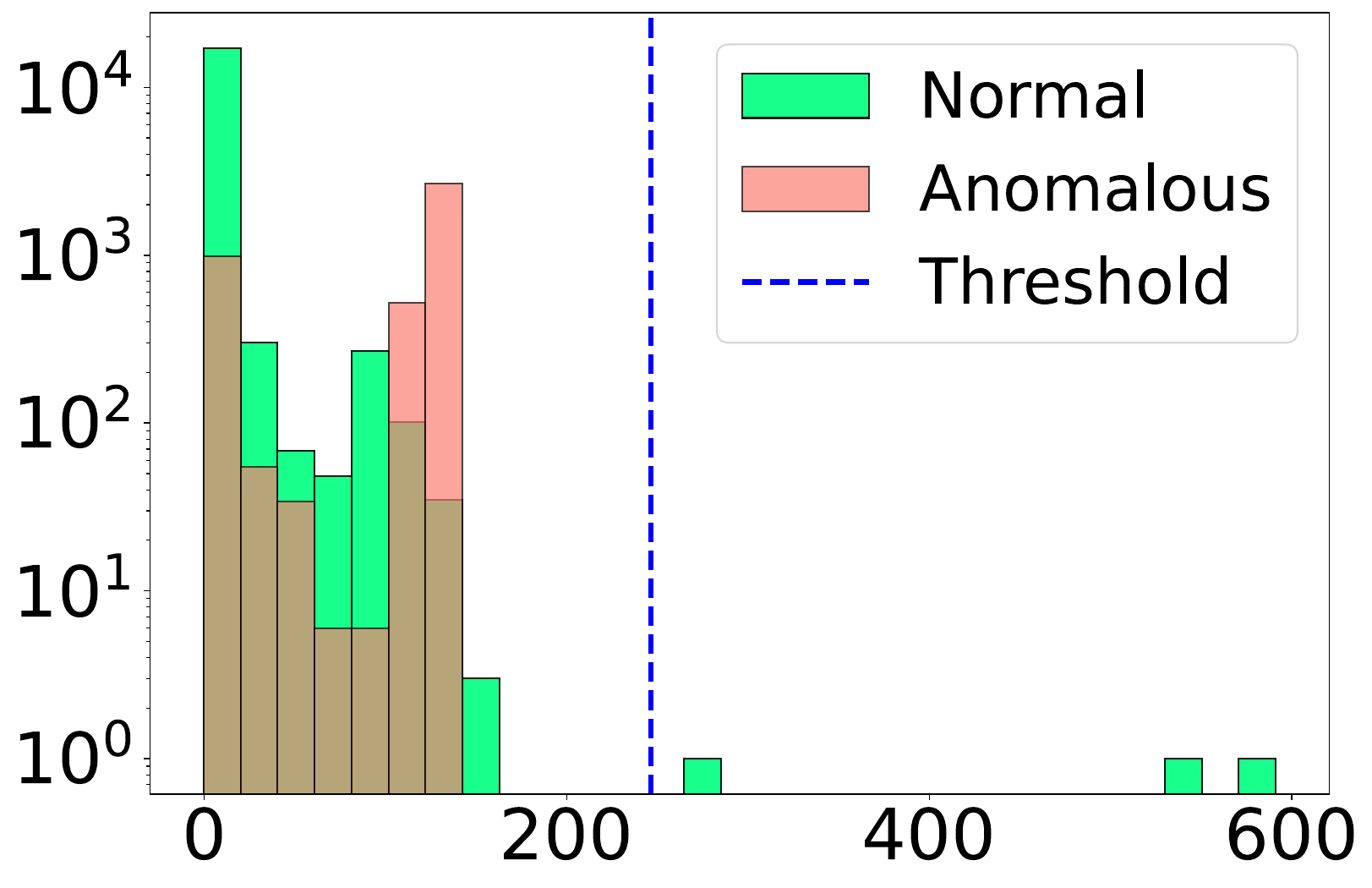}
    \end{minipage}
    \hfill
    \begin{minipage}[b]{0.49\linewidth}
    \centering
    \scriptsize{GRU}
        \includegraphics[width=\linewidth]{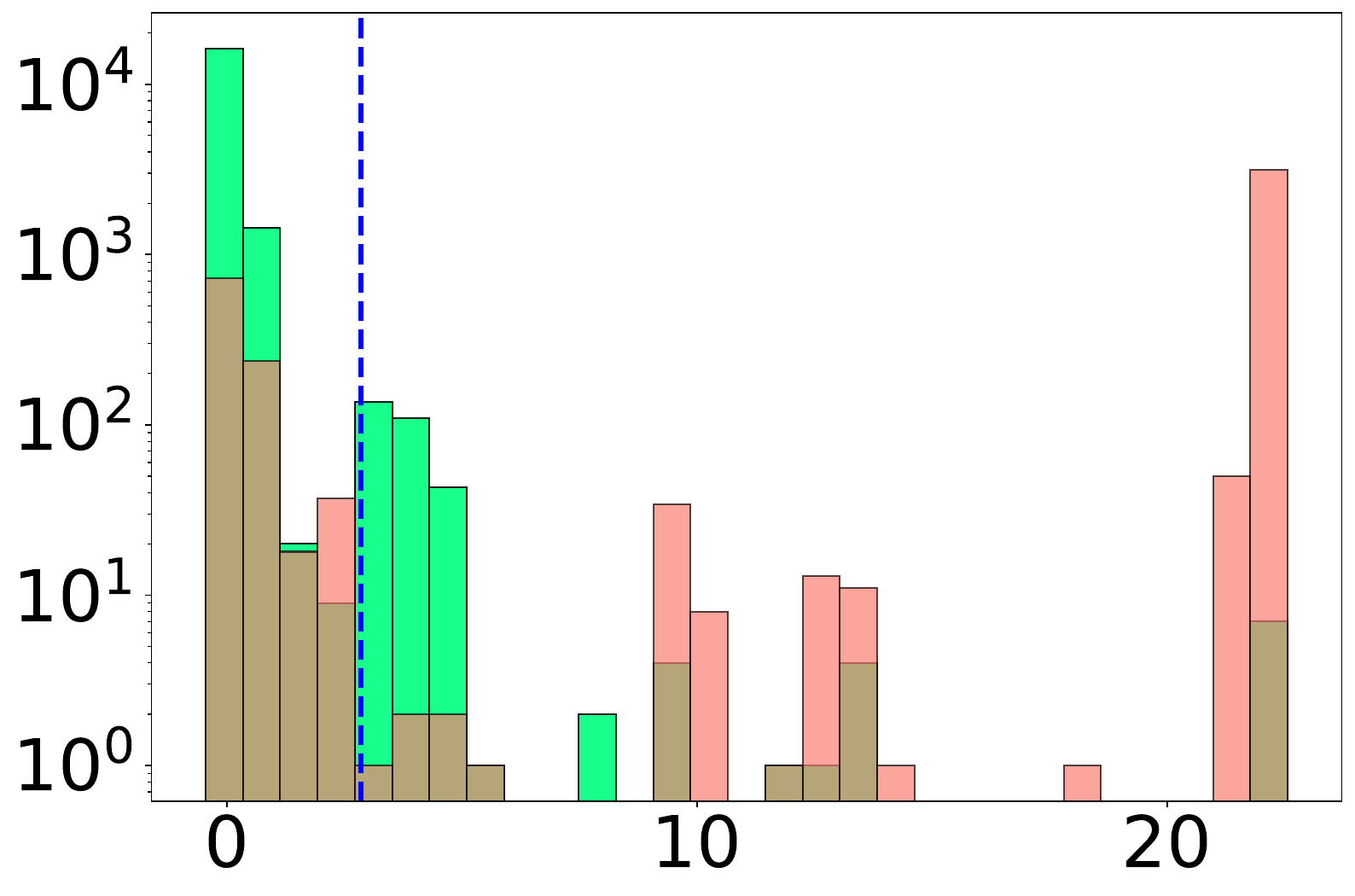}
    \end{minipage}
    \caption{
        Anomaly score distributions on the SWaT test set for all models (in logarithmic scale). 
        GRU and GDN exhibit better separation between normal (green bars) and anomalous (red) scores, while GCN and MTAD-GAT exhibit significant overlap, complicating threshold selection.
    }
    \label{fig:score_histograms_swat}
\end{figure}

The score distributions in Fig.\ \ref{fig:score_histograms_swat} further illustrate these challenges. Each histogram shows the normal (green) and anomalous (red) scores in the test set for all four methods. GRU and GDN produce relatively well-separated distributions, which helps explain their stronger and more consistent results. In contrast, GCN and MTAD-GAT exhibit substantial overlap between normal and anomalous scores, making reliable threshold selection considerably harder. For MTAD-GAT in particular, the chosen threshold (dashed vertical line) is clearly suboptimal for the test data; an outcome of the distribution shift between validation and test sets. When analyzing the same plot in the TELCO dataset, we observe that the score histograms are not well separated and lack a bimodal structure, which aligns with the lower VUS results reported earlier.

These experiments illustrate a downside of deriving anomaly scores from reconstruction or prediction losses, used as proxies for detecting abnormal behavior, particularly in terms of the threshold selection. 
While some evaluation metrics such as VUS are threshold-agnostic, real-world applications still require binary decisions, making threshold selection a critical step. 
We just shown that poor detection performance often stems not from the thresholding method itself (which may even be adaptive), but from the non-discriminative nature of the score distributions produced by certain models.
These findings highlight the limitations of proxy-based scoring and point toward the need for more task-aligned objectives, such as learning inherently discriminative representations through, for instance, contrastive learning.

\subsection{Impact of Graph Topology} \label{topology}
We now assess whether the incorporation of a graph structure improves performance by comparing the GCN and GDN models on various topologies: a fully connected graph (as before), the predefined or learned graph (SWaT and GDN only respectively), and finally a statistically inferred graph using the popular Meinshausen-Bühlmann (MB) method~\cite{Meinshausen_2006}.

In the SWaT dataset (see the upper portion of Table~\ref{tab:metricas_grafo_swat_es}), which features an underlying physical structure, employing an informative graph topology significantly improves performance, particularly in the GCN case. Note that in the GDN case, although the system topology obtains the best overall results, its attention mechanism makes it robust to the topology's choice. Furthermore, and quiet interestingly, GCN's best results are obtained when using the MB graph and not the system topology. The relationship between variables is thus better captured by the MB method. 



\begin{table}
\centering
\caption{VUS metrics (ROC and PR) for different graph topologies in the SWaT and TELCO datasets using the GDN and GCN models. The highest value is shown in \textbf{bold} and the second highest is \underline{underlined} for each model.}
\resizebox{\columnwidth}{!}{%
\begin{tabular}{@{}|l|l|l|c|c|@{}}
\toprule
\textbf{Dataset} & \textbf{Model} & \textbf{Graph Topology} & \textbf{$VUS$-$ROC$} & \textbf{$VUS$-$PR$} \\
\midrule
\multirow{10}{*}{SWaT} & \multirow{4}{*}{GCN} & Fully Connected  & 0.72             & 0.55             \\
                       &                      & System Topology  & 0.79             & 0.53             \\
                       &                      & MB               & \textbf{0.87}    & \textbf{0.76}    \\
                       &                      & Random Graph     & \underline{0.82} & \underline{0.63} \\
\cmidrule(l){2-5}
                       & \multirow{5}{*}{GDN} & Fully Connected  & 0.82             & 0.70             \\
                       &                      & GDN Graph        & \underline{\textbf{0.85}} & \underline{0.73} \\
                       &                      & System Topology  & \underline{\textbf{0.85}} & \textbf{0.75}    \\
                       &                      & MB               & 0.83             & 0.71             \\
                       &                      & Random Graph     & 0.83             & 0.70             \\
\midrule
\multirow{8}{*}{TELCO} & \multirow{3}{*}{GCN} & Fully Connected  & \underline{\textbf{0.64}} & \underline{\textbf{0.05}} \\
                       &                      & MB               & 0.62             & 0.04             \\
                       &                      & Random Graph     & \underline{\textbf{0.64}} & \underline{\textbf{0.05}} \\
\cmidrule(l){2-5}
                       & \multirow{4}{*}{GDN} & Fully Connected  & 0.60             & \underline{0.08} \\
                       &                      & GDN Graph        & \underline{0.65} & \underline{0.08} \\
                       &                      & MB               & 0.64             & 0.05             \\
                       &                      & Random Graph     & \textbf{0.67}    & \textbf{0.09}    \\
\bottomrule
\end{tabular}%
}
\label{tab:metricas_grafo_swat_es}
\end{table}

On the other hand, in the TELCO dataset no consistent improvement was observed when using different topologies, the best performance being achieved by a random graph. This indicates that using better graph inference  methods could lead to improved results, but it is not clear when the dataset does not have an explicit graph structure.


\subsection{Metric-Loss correlation}


As we discussed before, in many TSAD models, training is performed using regression objectives (e.g., forecasting or reconstruction loss), while evaluation relies on classification metrics. This raises a fundamental question: does minimizing regression loss improve anomaly detection performance?

To explore this, we analyzed the correlation between validation loss, calculated over normal data, and evaluation metrics, computed over the full validation set, including anomalies, across 200 trials from hyperparameter tuning of GCN, GDN, and GRU models on the SWaT dataset. Pearson correlation was used to measure linear relationships, with an ideal scenario corresponding to strong negative correlation (i.e., lower loss leads to better metric scores).

\begin{figure}
    \centering
    \includegraphics[width=\linewidth]{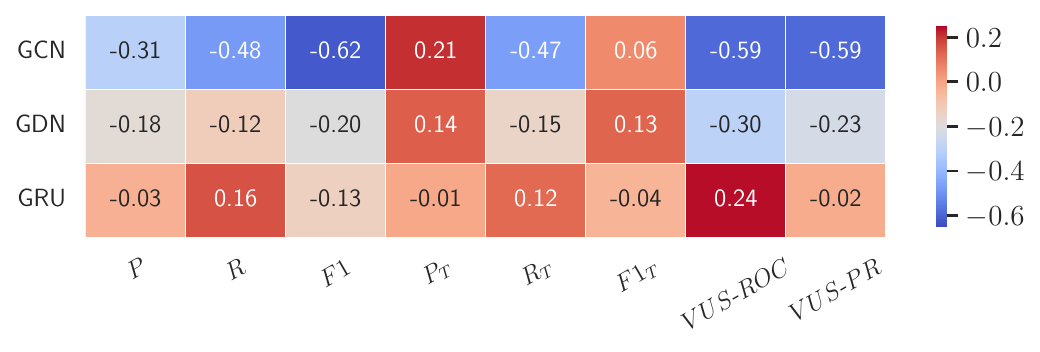}
    \caption{Correlation between the different metrics and the validation loss for the different models.}
    \label{fig:loss_metric_correlation_matrix}
\end{figure}

Figure~\ref{fig:loss_metric_correlation_matrix} shows the results. The GCN model, which performs worst, exhibits the strongest negative correlation between loss and VUS, suggesting that better forecasting results in an improved anomaly detection. In contrast, the GDN and GRU models achieve superior metric scores but show weak or even positive correlation, implying that a better regression fit does not translate to a better detection performance.

These findings suggest that optimizing purely for regression loss may be suboptimal. Alternative approaches, such as contrastive learning~\cite{CoLA2022,zahra2024carla}, which leverage anomaly labels during training to structure the feature space more effectively, could offer a more aligned and robust solution for anomaly detection tasks.

\subsection{Interpretability analysis}

Beyond accurate detection, anomaly detection models should help identify where anomalies originate. Graph-based models, especially GDN with attention mechanisms, provide a natural framework for this by modeling sensor dependencies and highlighting influential nodes.

To assess interpretability, we analyze each model's ability to attribute detected anomalies to specific sensors in the SWaT dataset, focusing on a known event affecting sensor FIT401. We compare GDN using both a learned and predefined SWaT topologies, to a the GRU baseline. For GDN, we further examine the distribution of attention weights to identify which sensors most strongly influence the anomaly score.

All models consistently rank the true anomalous sensor among the top sensors during the anomaly period. This indicates that, while score-based methods can suggest likely affected sensors, interpretability still requires further analysis. 

\begin{figure}
\centering
\includegraphics[width=0.7\linewidth]{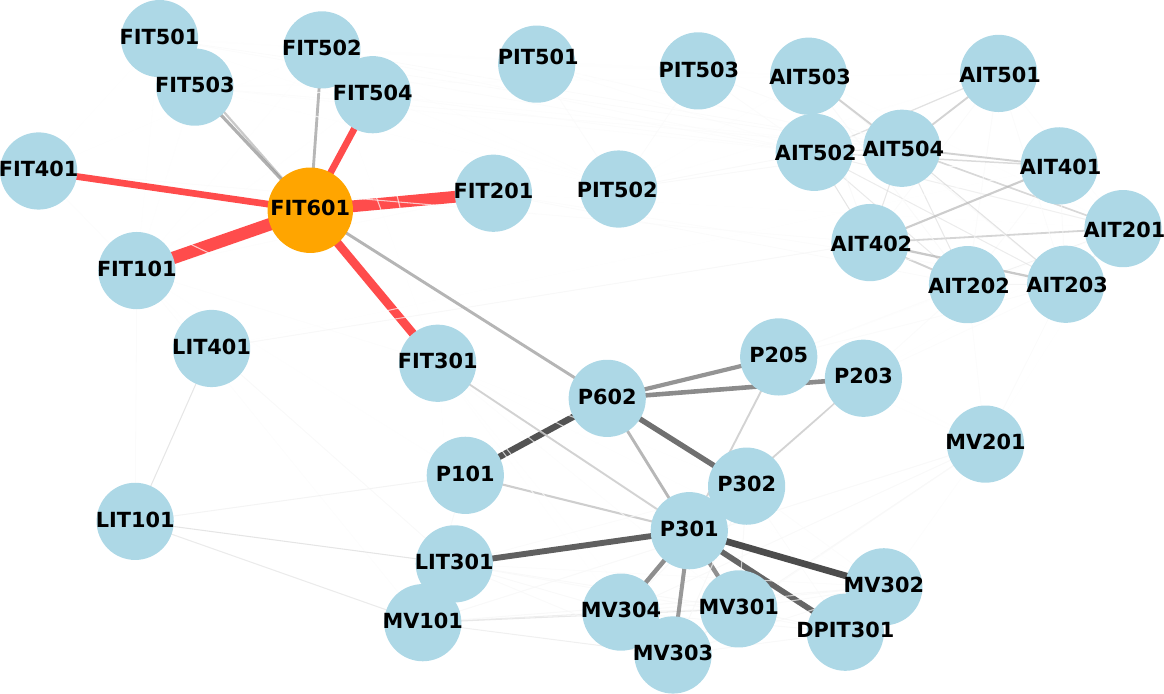}
\caption{Visualization of attention during the anomaly in FIT401 using the SWaT topology. The node with the highest anomaly score is shown in orange; red edges indicate the five neighbors with the highest attention weights. Attention concentrates on physically connected nodes, improving interpretability and mapping directly to system flow: an anomaly in a FIT sensor.}
\label{fig:attention_graphs}
\end{figure}

In particular, attention visualization for GDN with the SWaT topology reveals more coherent and physically meaningful patterns. Figure~\ref{fig:attention_graphs} shows attention distributions during the anomaly. The node with the highest anomaly score corresponds to FIT601, while the actual anomaly occurs in FIT401. However, the strongest attention edges are concentrated among FIT sensors (FIT401, FIT101, FIT201), all of which measure related physical quantities. This indicates that the model correctly focuses on a group of related sensors, improving interpretability and helping to identify a consistent set of potentially anomalous sensors. When other graph structures are used, attention becomes dispersed across unrelated nodes, reducing interpretability.

\begin{figure}
    \centering
    \begin{minipage}[b]{0.38\linewidth}
    \centering
    \scriptsize{GRU - PIT501}
    \includegraphics[width=\linewidth]{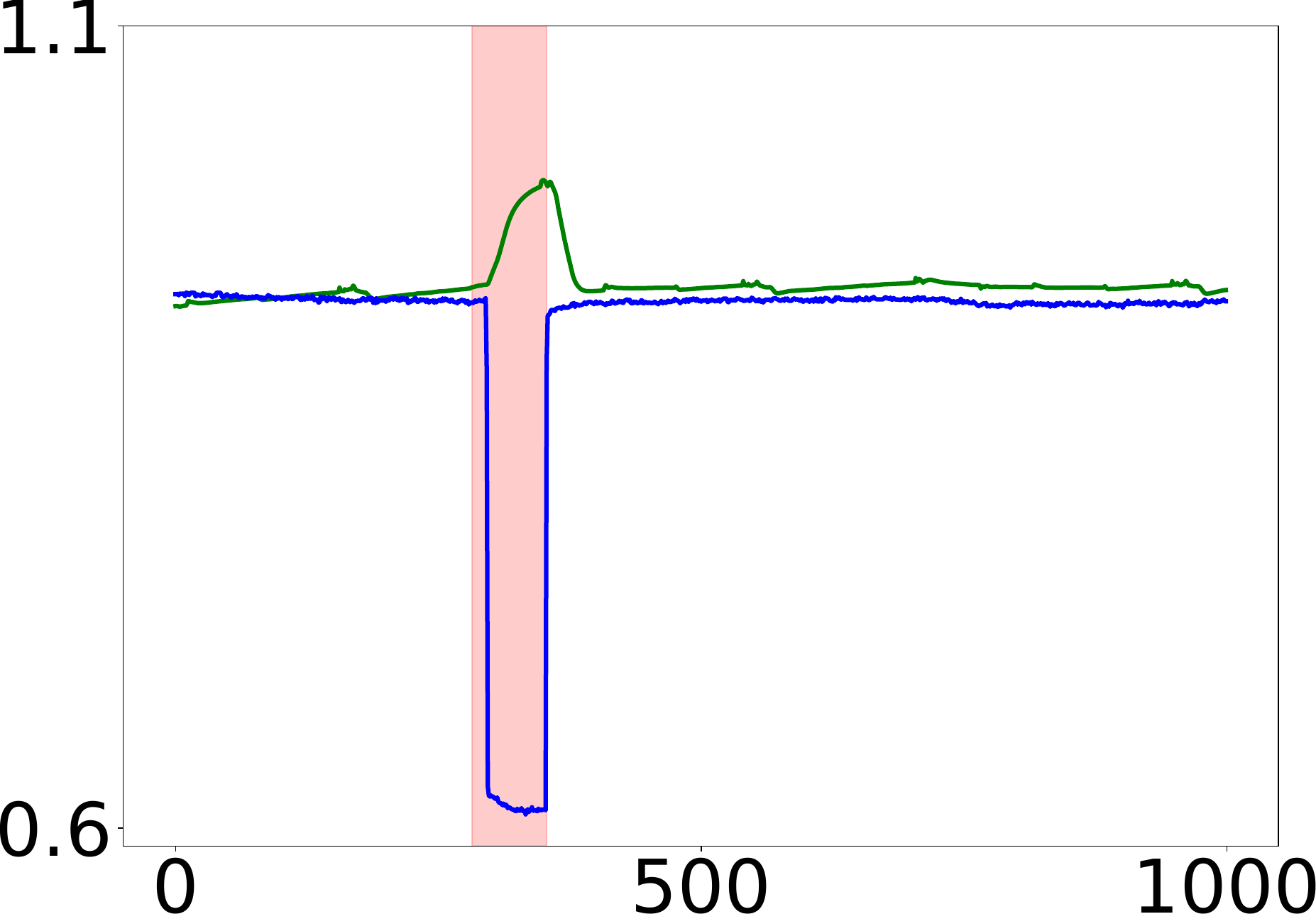}
    \end{minipage}
    \begin{minipage}[b]{0.38\linewidth}
    \centering
    \scriptsize{GDN - PIT501}
    \includegraphics[width=\linewidth]{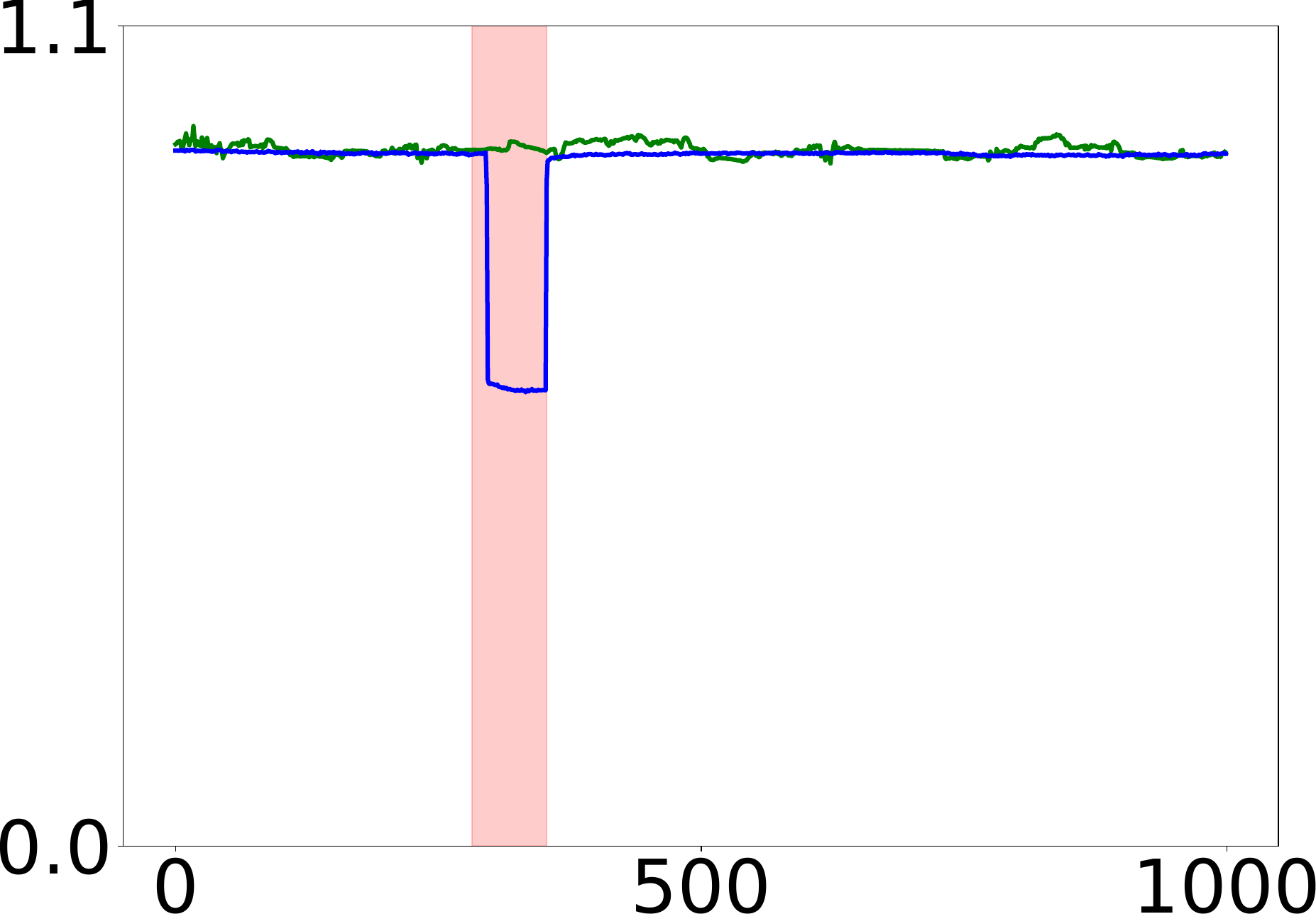}
    \end{minipage}
    \begin{minipage}[b]{0.38\linewidth}
    \centering
    \scriptsize{GRU - P203}
    \includegraphics[width=\linewidth]{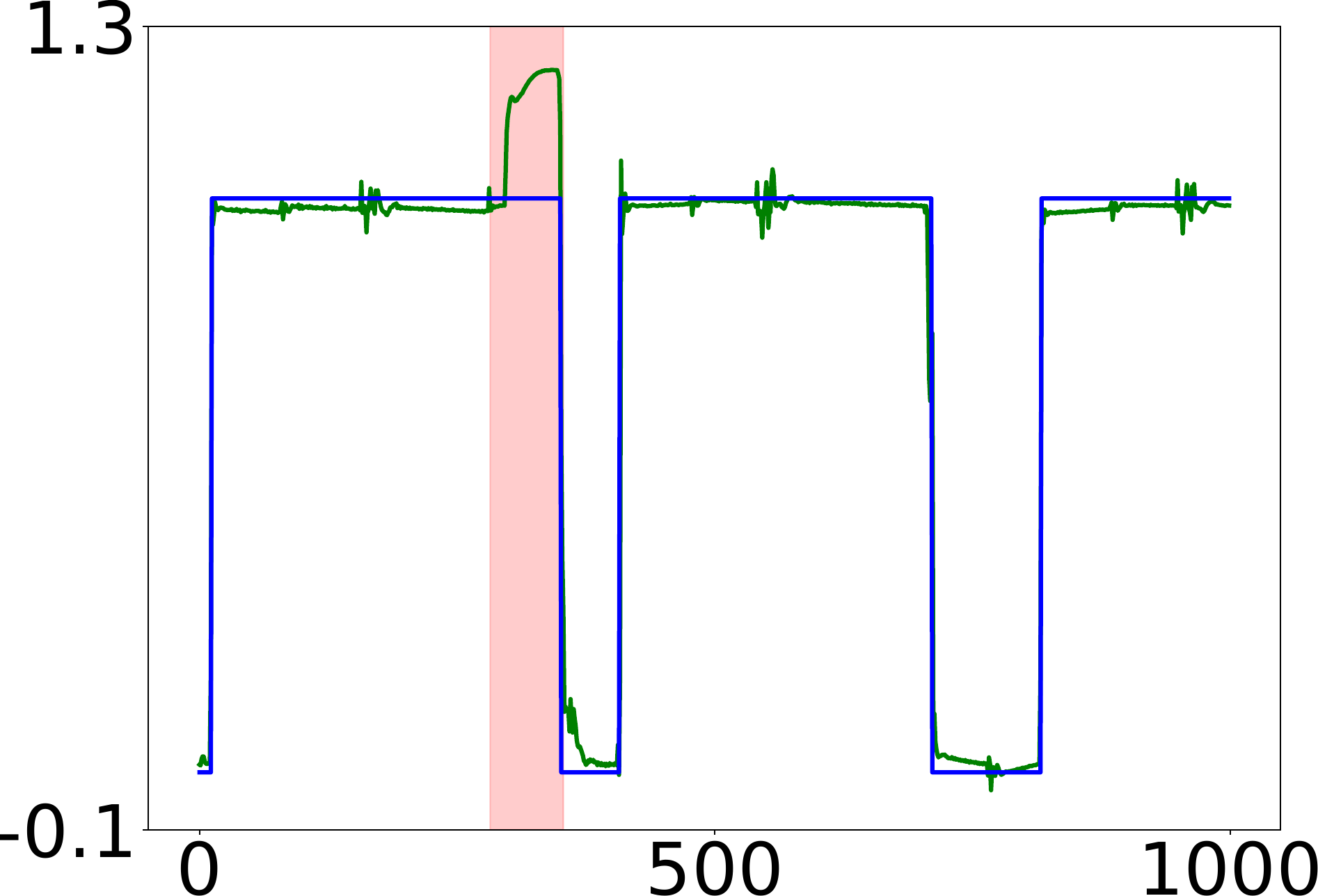}
    \end{minipage}
    \begin{minipage}[b]{0.38\linewidth}
    \centering
    \scriptsize{GDN - P203}
    \includegraphics[width=\linewidth]{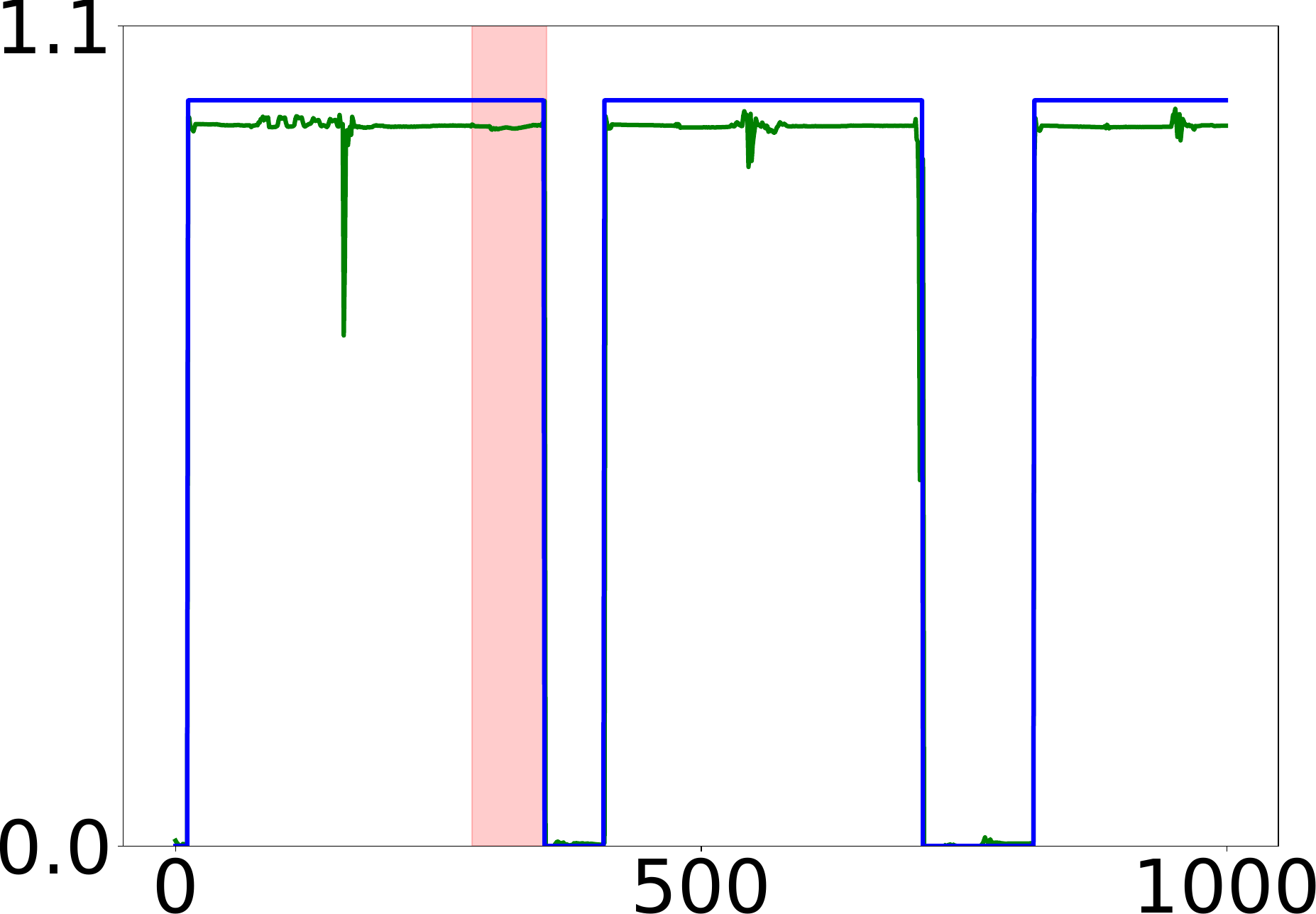}
    \end{minipage}
    \caption{Forecast comparison between GDN with graph topology (right) and GRU (left) on the SWaT dataset. The blue line represents the true values of the time series, the green line shows the forecasted values, and the red shaded regions indicate anomaly labels. GDN's use of a meaningful system topology results in stable forecasts and clear, localized anomaly detection. In contrast, GRU lacks this structure, leading to unstable forecasts and reduced interpretability, as anomalies can affect all sensors. }
    \label{fig:gru_gdn_forecast_comparison}
\end{figure}

Using a graph topology not only improves attention distributions, but also stabilizes prediction scores, as exemplified in Fig.~\ref{fig:gru_gdn_forecast_comparison}. Here, GDN’s graph-based approach keeps forecasts stable and restricts anomaly effects to the affected sensor (PIT501 in stage 5 of the process in this example), making fault localization straightforward. In contrast, GRU forecasts are less stable; an anomaly in stage 5 also deteriorates predictions for stage 2 (P203), making it difficult to pinpoint the true source of the anomaly.

Thus, predefined topologies enhance interpretability by aligning attention with real system structure and stabilizing the predictions of the considered models.





\section{Conclusions}\label{sec:conclusions}

In this work, we introduced a modular and open-source framework for evaluating graph-based models in TSAD. This framework enables reproducible experimentation across datasets, architectures, graph topologies, and evaluation metrics. Using it, we conducted a comparative study of several GNN-based methods and baselines across two real-world datasets with differing structural characteristics.

Our findings show that GNNs can provide competitive, and in some cases superior, performance in TSAD tasks, particularly when there is an underlying and explicit graph. More importantly, they offer improved interpretability by localizing anomalies to specific nodes in the input graph. We also found that attention-based GNNs are more robust to uncertainty in graph construction, making them attractive for use in semi-structured or anonymized datasets.
Alongside model evaluation, we critically examined the limitations of commonly used performance metrics and scoring strategies. In particular, we highlighted how score distributions and threshold sensitivity can undermine the reliability of evaluation.

Looking ahead, our findings suggest that moving beyond proxy-based scoring (e.g., reconstruction error) could further improve TSAD systems. In this context, contrastive learning offers a promising direction for producing more discriminative anomaly scores directly aligned with the detection task~\cite{CoLA2022,zahra2024carla}, which we plan to explore and integrate to our framework in future work.

\bibliographystyle{apalike}
{\small
\bibliography{referencias}}

\end{document}